\pdfoutput=1

\documentclass[11pt]{article}

\usepackage[final]{acl}

\usepackage{times}
\usepackage{latexsym}

\usepackage[T1]{fontenc}

\usepackage[utf8]{inputenc}

\usepackage{microtype}

\usepackage{inconsolata}

\usepackage{graphicx}
\graphicspath{./images}

\usepackage{booktabs}
\usepackage{multirow}
\usepackage{amsmath}

\title{Data-Constrained Synthesis of Training Data for De-Identification}

\author{Thomas Vakili, Aron Henriksson, and Hercules Dalianis \\
Department of Computer and Systems Sciences\\ Stockholm University,  Kista, Sweden\\ 
\texttt{\{thomas.vakili, aronhen, hercules\}@dsv.su.se}
}

\begin{document}
\maketitle
\begin{abstract}

Many sensitive domains -- such as the clinical domain -- lack widely available datasets due to privacy risks. The increasing generative capabilities of large language models (LLMs) have made \textit{synthetic datasets} a viable path forward. In this study, we domain-adapt LLMs to the clinical domain and generate synthetic clinical texts that are machine-annotated with tags for personally identifiable information using capable encoder-based NER models. The synthetic corpora are then used to train synthetic NER models.
The results show that training NER models using synthetic corpora incurs only a small drop in predictive performance. The limits of this process are investigated in a systematic ablation study -- using both Swedish and Spanish data. Our analysis shows that smaller datasets can be sufficient for domain-adapting LLMs for data synthesis. Instead, the effectiveness of this process is almost entirely contingent on the performance of the machine-annotating NER models trained using the original data. %
\end{abstract}

\section{Introduction}

Many useful applications in NLP involve domains where the data are sensitive. These privacy risks, and the accompanying limits to sharing data, have traditionally been solved through \emph{de-identification}. This process involves finding parts of the text that can be used to identify an individual. Such information is typically referred to as personally identifiable information (PII). After locating PII, the passages need to be processed to remove or obscure the PII. Traditionally, this time-consuming work has been done manually. Automatic de-identification \cite{meystre2010automatic} is a machine-driven approach that typically relies on named entity recognition (NER) to detect PII that needs to be removed.

Unfortunately, the PII datasets that exist to assist in privacy preservation are themselves sensitive and can usually not be shared. This circularity, together with the increasing generative capabilities of large language models (LLMs), has led to a growing interest in overcoming data limitations by eschewing the use of real data altogether. Instead, one can use generated \textit{synthetic} corpora.

Previous studies have mainly been concerned with evaluating the privacy of the synthetic text \cite{Yue-etal-2023,Miranda-etal-2024} or with creating the strongest-performing model possible using synthetic data \cite{libbi2021generating,hiebel2023can,Liu-etal-2025}. Our study instead examines how synthetic data can be produced under constrained resources. This understudied problem is common in clinical institutions that lack resources, both in terms of data and computational hardware.

We carry out a systematic evaluation of key factors impacting the utility of LLM-generated synthetic data as training data for downstream tasks. Specifically, we study synthetic NER data for PII detection -- an important task 
in the privacy-sensitive %
healthcare domain. Synthetic clinical data are generated using domain-adapted LLMs and machine-annotated using fine-tuned NER models. The evaluations focus primarily on the \emph{utility} of synthetic data for training NER models.%

Through extensive experimentation, we investigate the impact on utility of (i) the amount of data used for domain adaptation of the synthesizing LLM, (ii) the quality of the machine annotator, (iii) the amount of synthetic data generated, and (iv) model size.  We also quantify the diversity and privacy of the generated data, and carry out experiments across two languages -- Swedish and Spanish. Our main contributions are:

\begin{enumerate}
    \item Demonstrating that moderately-sized LLMs can be adapted to the clinical domain to produce high-utility text with relatively small amounts of in-domain data.
    \item Showing that using synthetic machine-annotated data allows for training NER models that perform only slightly worse compared to using real, sensitive data, while reducing the risk of exposing sensitive information in the original data.
    \item Finding that, for the task of detecting PII, using larger generative LLMs for synthesis does not yield clear improvements in terms of utility. Rather, downstream performance relies on having a high-quality gold standard NER model for providing machine annotations.
\end{enumerate}

\section{Related Research}
There have been several approaches to generating synthetic clinical data for a number of languages and for different purposes. Broadly speaking, most prior works have either focused on maximizing the utility of the synthetic data, or on studying the privacy characteristics of synthetic corpora.

While most papers studying data synthesis contain some form of privacy analysis, other papers have this as their main focus. Several papers study how differentially private learning impacts the utility \cite{Yue-etal-2023,igamberdiev_dp-nmt_2024} and the privacy of the data \cite{Miranda-etal-2024}. While privacy is an important justification for synthesizing data, it is not the main focus of our paper.

The second main current in the literature explores how to optimize synthesis to create the best possible synthetic corpora. These papers synthesize data using locally domain-adapted LLMs \cite{ive2020generation,hiebel2023can}, or using instruction-tuned models \cite{kiefer2024instruction,Liu-etal-2025}. They show that high-utility data synthesis is possible. However, fewer papers systematically examine the conditions required for success.

In our literature review, two studies stand out as particularly relevant to this study. \citet{libbi2021generating} synthesize a Dutch corpus for PII detection using a GPT-2 model \cite{radford_language_2019} domain-adapted using 1 million documents and add rule-based machine annotations. Our study follows the same overall process for synthesis, but uses much less data and more modern NLP techniques. \citet{Xu-etal-2023a} similarly create synthetic corpora and experiment with constraining the total amount of data used, but do so for the relation extraction task. In this paper, we focus on a different task -- NER for PII detection. Furthermore, in contrast to both studies, we systematically evaluate the impact of constraining data alternately for \textit{both} domain adaptation \textit{and} machine annotation, try two different model sizes, synthesize corpora of different sizes, and validate our results across two languages.

\section{Data and Methods}

\begin{figure*}[ht!]
    \centering
    \includegraphics[width=\textwidth]{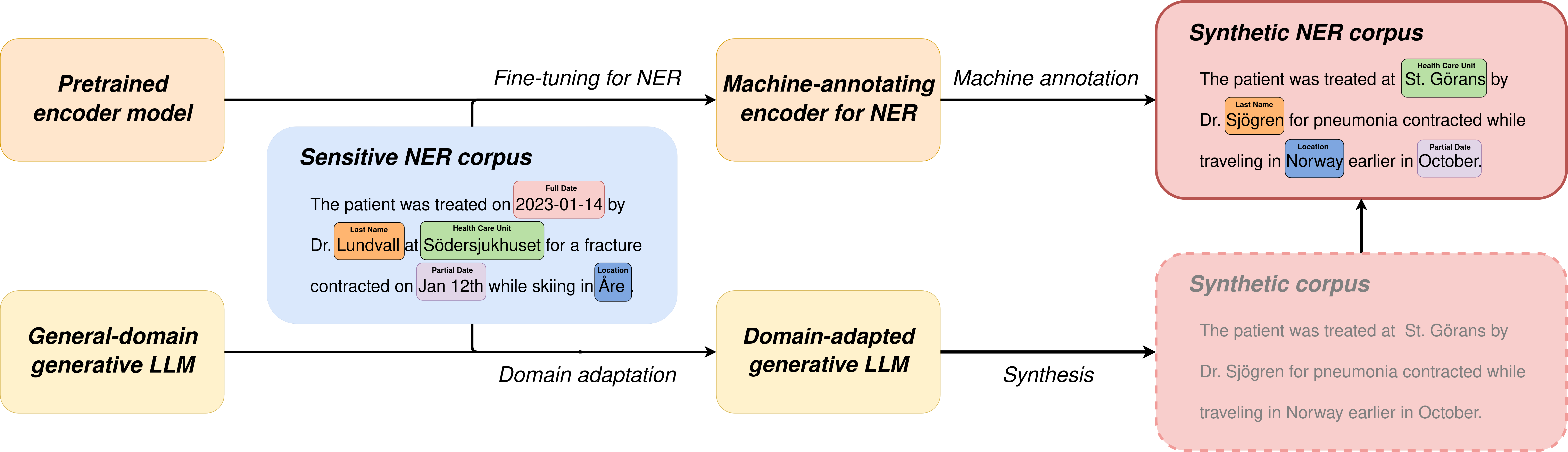}%
    \caption{%
    This figure illustrates the steps for creating the synthetic corpora. A gold standard corpus is used to train a NER model and to domain-adapt a general-domain LLM to produce synthetic clinical data. The LLM is used to generate synthetic data, and the NER model is used to add machine annotations. Later, these corpora are used to train synthetic NER models that are evaluated on gold-standard test data.}
    \label{fig:process}
\end{figure*}

In this study, we investigate the impact of various factors related to generating synthetic data for fine-tuning encoder language models on downstream tasks in the healthcare domain. Specifically, we study the possibility of generating synthetic clinical text for training NER models for detecting PII. The synthetic text is created by a domain-adapted generative LLM and then machine-annotated for PII using a fine-tuned encoder model. This process follows the structure of previous works \cite{libbi2021generating} and is illustrated in Figure \ref{fig:process}.%

\subsection{Generative Models}
The aim of this study is to examine the feasibility of generating training data for NER models detecting PII. The foundation of the training data are synthetic texts, generated using autoregressive LLMs. Two model families are used as a base for domain adaptation to the clinical domain.

\begin{description}
    \item[GPT-SW3]
    For Swedish, we use the \mbox{\textit{GPT-SW3}} model \cite{Ekgren-etal-2024}. This autoregressive language model was trained using approximately 320 billion tokens. The data were mainly composed of Scandinavian texts and 35.3\% of the data is Swedish.
    
    \item[FLOR]
    The autoregressive model used to generate Spanish data is the \textit{FLOR} model \cite{DaDalt-etal-2024}. The model was initialized with the weights of the multi-lingual BLOOM model \cite{Scao-etal-2023} and trained with continued pre-training. The data used spanned 140 billion tokens and was composed of equal parts English, Spanish, and Catalan data.
\end{description}

Both models are used autoregressively, without instruction tuning. The hypothetical -- but often occurring -- scenario motivating the study's design is where researchers have access to a small and sensitive NER dataset that cannot be shared outside of their organization. Zero-shot synthesis is an alternative strategy, but we leave it to future research to evaluate if this approach can yield clinical texts that are sufficiently similar to the real data. Instead, we perform different degrees of domain-adaptive fine-tuning to train the LLMs to produce such texts.

The primary experiments in Section \ref{sec:eval-amount} used the versions of FLOR and GPT-SW3 with 6.3 billion and 6.7 billion parameters, respectively. Smaller models with 1.3 billion parameters were used in Section \ref{sec:model-size} to investigate the impact of using smaller %
models for synthesis. %

\subsection{NER Datasets}
\label{sec:datasets}

This study focuses on a particular type of NER dataset -- NER for detecting PII. Such datasets exist for several languages but are, as discussed in the introduction, very difficult to share and access. This is particularly true for datasets targeting the clinical domain. In this study, two datasets for detecting PII in clinical data are used.%

\begin{description}
     \item[SEPR PHI] 
The \textit{Stockholm EPR PHI\footnote{Protected health information (PHI) is a term defined by the HIPAA regulation \citep{HHS-1996}. The PHI term covers a subset of the types of information that constitute PII.} Pseudo Corpus (SEPR PHI)} is a Swedish dataset from five different healthcare units consisting of 100 patient records split into 21,553 sentences. The corpus spans 282,766 tokens, where 6,755 are manually annotated for nine different PII classes \cite{velupillai2009developing}. The dataset has then been pseudonymized, meaning that all the annotated entities have been replaced with realistic pseudonyms \cite{dalianis2019pseudonymisation}.

    \item[MEDDOCAN]
The second dataset is \textit{MEDDOCAN} -- a Spanish dataset consisting of 1,000 medical texts. These are based on clinical cases augmented with PII from auxiliary sources \cite{Marimon-etal-2019}. The texts were then manually annotated for 19 different PII. Out of 504,569 tokens,  41,859 are tagged as PII. The documents were originally divided into train, test, and development sets. In this paper, these three subsets have been merged before resampling them into new subsets for five-fold cross-validation.
\end{description}

Each dataset was split into three subsets: one for training, another for validation, and a third held-out subset for testing. The training sets are used to domain-adapt the generative models and to train the machine-annotating NER models. The purpose of the validation subsets is twofold. First, they are used to monitor the training processes when fine-tuning the models for synthesis and NER. Once the models for synthesis have been trained, the validation data also serve as starting points when creating the synthetic corpora. Finally, the quality of the NER models trained using the synthetic corpora is evaluated using the held-out test sets, as these are not in any way part of the training or synthesis processes.

\subsection{Encoder Models}
\label{sec:method-encoders}
This study trains NER models by fine-tuning pre-trained encoder models. Two different models are used, one for each language.

\begin{description}
    \item[SweDeClin-BERT]
We use \textit{SweDeClin-BERT} \cite{Vakili-etal-2022} for Swedish data as it has previously shown strong performance on the SEPR PHI corpus. This BERT-style model is based on the general-domain KB-BERT model \cite{Devlin-etal-2019,Malmsten-etal-2020} and adapted to the clinical domain through continued pre-training on the Swedish Health Bank corpus \cite{Dalianis-etal-2015}. It consists of 125 million parameters.
    
    \item[roberta-base-bne]
For Spanish data, we use the \textit{roberta-base-bne} model trained by \citet{Gutierrez-Fandino-etal-2022}. This RoBERTa-based model \cite{liu_roberta_2019} consists of 125 million parameters. It was trained using a large Spanish corpus collected by the National Library of Spain and performs strongly on the MEDDOCAN task.
\end{description}

These models are used for two purposes. First, they are fine-tuned using the gold standard datasets. These gold models are used to add machine annotations to the synthesized corpora. They are also used as baselines for the later, synthetic NER models. The synthetic models are also initialized from the pre-trained encoder models, but are fine-tuned using the synthetic, machine-annotated corpora. The gold and synthetic NER models are then evaluated and compared to each other in order to measure the performance implications of using synthetic data.

\subsection{Synthesizing and Evaluating the Corpora}
\label{sec:method-synth}
The generative models were domain-adapted by fine-tuning them for autoregressive language modeling using QLoRA \cite{Dettmers-etal-2023} as implemented in the Axolotl framework\footnote{\url{https://github.com/axolotl-ai-cloud/axolotl}} with $r=8$ and $\alpha=32$. %
A comprehensive specification of all hyperparameters is available in Appendix \ref{appendix:ft-params}. %
GPT-SW3 was domain adapted using the Swedish SEPR PHI corpus and FLOR using the Spanish MEDDOCAN corpus. Domain adaptation was carried out with varying amounts of data to determine the impact on the utility of the synthetic data.

As mentioned in Section \ref{sec:datasets}, the validation sets used for monitoring the fine-tuning process were also used for data synthesis. The 5\% validation subsets were used to create starting points for generating text, as suggested by \citet{libbi2021generating}. The starting points were created by taking the first three words of each document in the validation sets.

Synthetic corpora have an intriguing advantage over real corpora: they can be arbitrarily large. Taking this feature into account, each three-word starting point was used to create 80 new samples. Consequently, the synthetic corpora are four times larger than the gold-standard corpora. In Section \ref{sec:synthesis-amount}, the benefits of exploiting this feature %
are examined through experiments that use smaller amounts of synthetic data. %

Synthesis was done using the vLLM library \cite{Kwon-etal-2023}. We use nucleus sampling \cite{Holtzman-etal-2020} with $p=0.95$, and the minimum token length is set to 10. The maximum token length is set to the length of the longest document in the validation set, or at least 50. The temperature was set to $t=1.0$ after preliminary experiments showed that varying $0.8 \leq t \leq 1.2$ had very little impact on the results.

Finally, the synthetic texts were machine-annotated for PII entities. These were added using NER models fine-tuned on the gold-standard datasets. The gold-standard and synthetic NER models were trained for 6 epochs with a batch size of 16. The limited context window of the models was overcome, both during training and machine annotation, by splitting long documents into 128-word chunks. This chunking was done both during training and when using the gold-standard models for machine annotation.

The whole process, from domain adaptation to training and evaluating the synthetic NER models, was done through five-fold cross-validation. The utility of the synthetic corpora was measured using the token-level F\textsubscript{1} score. These evaluations rely on each fold's held-out gold-standard test data.

\section{Experiments}
\label{sec:experiments}

The main contribution of this paper is a systematic investigation into how the different steps of synthetic corpora creation respond to data constraints. This section describes these experiments and their results. All experiments, except for those in Section \ref{sec:synthesis-amount}, take advantage of the unbounded nature of synthetic corpora and allow them to be four times larger than the gold standard datasets. The performance of the NER models is summarized using token-level F\textsubscript{1} scores tested on gold-standard data. All values are averages and standard deviations that are calculated based on the results from the five-fold cross-validation.

\subsection{Constraining the Total Amount of Data}
\label{sec:eval-amount}
\begin{table*}[ht]
    \centering
    \resizebox{\textwidth}{!}{
    \begin{tabular}{c c c c c c c c}
         \toprule
         \multirow{2}{*}{\textbf{\% of fold }} & \multicolumn{3}{c}{\textbf{SEPR PHI}} & & \multicolumn{3}{c}{\textbf{MEDDOCAN}}  \\
         \cmidrule{2-4} \cmidrule{6-8}
         & \textbf{Gold} & \textbf{Synthetic} & \textbf{$\Delta$} &
         & \textbf{Gold} & \textbf{Synthetic} & \textbf{$\Delta$} \\
         \midrule
\textit{5\%} &    0.707 $\pm$ 0.037 &     0.724 $\pm$ 0.035 &     -0.017 $\pm$ 0.051 &
& 0.931 $\pm$ 0.012 & 0.309 $\pm$ 0.060 &    0.622 $\pm$ 0.061 \\
\textit{25\%} &   0.871 $\pm$ 0.010 &     0.847 $\pm$ 0.010 &     0.024 $\pm$ 0.014 &
& 0.967 $\pm$ 0.003 & 0.964 $\pm$ 0.005 &     0.003 $\pm$ 0.006 \\
\textit{50\%} &   0.908 $\pm$ 0.007 &     0.885 $\pm$ 0.010 &     0.023 $\pm$ 0.012 &
& 0.973 $\pm$ 0.004 & 0.970 $\pm$ 0.004 &     0.003 $\pm$ 0.006 \\
\textit{95\%} &   0.926 $\pm$ 0.005 &     0.896 $\pm$ 0.007 &     0.029 $\pm$ 0.009 &
& 0.978 $\pm$ 0.005 & 0.973 $\pm$ 0.003 &     0.005 $\pm$ 0.006 \\
         \bottomrule
    \end{tabular}
    }
    \caption{Between 5\% and 95\% of the training data in each fold was used to domain-adapt the generative LLMs and the machine-annotating encoder models. In this table, the F\textsubscript{1} scores are listed for both the NER models trained on gold standard data and the synthetic data, as well as the difference. The F\textsubscript{1} scores are the average scores and their standard deviation over all five folds.}
    \label{tab:data-4-both-steps}
\end{table*}

The first experiment of this study investigated how much data is required to produce a well-performing NER model for detecting PII. The amount of data used for domain-adapting the generative model and fine-tuning the gold NER model is varied. This models the common situation where there is limited access to data, and demonstrates what performance can be expected for different data sizes.

Within each fold, this is scaled to between 5\% and 95\% of the training data in the fold. Four different amounts are used: 5\%, 25\%, 50\%, and 95\%. These subsets correspond to the \textit{Sensitive NER corpus} in Figure \ref{fig:process}. The final 5\% are used for validation and for creating prompts for synthesis.

Synthetic corpora have an advantage over real corpora: their size is only constrained by the amount of computing power available for generation. As explained in Section \ref{sec:method-synth}, this feature is incorporated by letting the synthetic data be four times larger than the original datasets. In later experiments, we analyze the extent to which this advantage helps. 

Table \ref{tab:data-4-both-steps} lists the average F\textsubscript{1} scores and their standard deviation for each tested configuration. Unsurprisingly, the performance of the gold models -- the models trained on the real data -- increases as more and more data are available for training. The MEDDOCAN models are more resilient to shrinking the amount of training data, but both cases show clear improvements as more data are available. The F\textsubscript{1} scores of the models trained on synthetic corpora follow a similar pattern. Increasing the data allows for more data to be used to domain-adapt the generative model and for training a better machine annotator.%

\subsection{Scaling the Data for Domain Adaptation}
The experiments in the previous section show that it is indeed feasible to create well-performing NER models trained on synthetic data using our method. On the other hand, the results depend on the amount of data used for domain-adapting the synthesizing generative LLM and for fine-tuning the machine-annotating encoder model. In the previous experiment, the data were fixed for both purposes.

This section describes an ablation study that measures the impact of varying the amounts of data used for domain adaptation. As before, the synthetic corpora are allowed to be four times larger than the original corpora. In these experiments, the amount of data used for fine-tuning the machine annotator is kept constant at 95\%, while the amount used for domain adaptation is varied between 5\% and 95\%. Additionally, we also %
use synthetic corpora generated \textit{without} domain adaptation.

\label{sec:data-4-domain-adaptation}
\begin{table}[t]
    \centering
    \resizebox{0.45\textwidth}{!}{
    \begin{tabular}{c c c}
    \toprule
    \textbf{\% for d.a.} & \textbf{SEPR PHI} & \textbf{MEDDOCAN} \\
    \midrule
\textit{0\%} &    0.547 $\pm$ 0.178 &     0.295 $\pm$ 0.011 \\
\textit{5\%} &    0.873 $\pm$ 0.014 &     0.313 $\pm$ 0.032 \\
\textit{25\%} &   0.877 $\pm$ 0.010 &    0.970 $\pm$ 0.005 \\
\textit{50\%} &   0.896 $\pm$ 0.007 &    0.970 $\pm$ 0.005 \\
\textit{95\%} &   0.896 $\pm$ 0.007 &    0.973 $\pm$ 0.003 \\
\midrule
\textit{Gold} & \textit{0.926 $\pm$ 0.005} & \textit{0.978 $\pm$ 0.005} \\
\bottomrule
    \end{tabular}
    }
    \caption{The amount of data used for \textbf{d}omain \textbf{a}daptation (d.a.) of the synthesizing generative LLM was varied from 0\% to 95\% of the training data in each fold. The average F\textsubscript{1} scores of the synthetic NER models and the gold-standard models are listed.}
    \label{tab:data-4-domain-adaptation}
\end{table}

The average F\textsubscript{1} scores of the models resulting from these experiments are listed in Table \ref{tab:data-4-domain-adaptation}. Unsurprisingly, the worst-performing models are those that were trained using corpora synthesized without domain adaptation. These results show that domain adaptation does matter. However, there are clearly diminishing returns from increasing the amount of data for domain adaptation. Increasing the amount of data from 50\% to 95\% produces nearly identical results.

\subsection{Varying the Data for Machine Annotation}
\label{sec:data-4-machine-annotation}
\begin{table*}[ht]
    \centering
    \resizebox{0.8\textwidth}{!}{
    \begin{tabular}{c c c c c c}
         \toprule
         \multirow{2}{*}{\textbf{\% for m.a. }} & \multicolumn{2}{c}{\textbf{SEPR PHI}} & & \multicolumn{2}{c}{\textbf{MEDDOCAN}}  \\
         \cmidrule{2-3} \cmidrule{5-6}
         & \textbf{Gold} & \textbf{Synthetic} &
         & \textbf{Gold} & \textbf{Synthetic} \\
         \midrule
        
\textit{5\%} &    0.707 $\pm$ 0.037 &       0.725 $\pm$ 0.039 &
&       0.931 $\pm$ 0.012 & 0.942 $\pm$ 0.010 \\
\textit{25\%} &   0.871 $\pm$ 0.010 &   0.858 $\pm$ 0.012 &
&       0.967 $\pm$ 0.003 & 0.967 $\pm$ 0.004 \\
\textit{50\%} &   0.908 $\pm$ 0.007 &   0.889 $\pm$ 0.005 &
&       0.973 $\pm$ 0.004 & 0.965 $\pm$ 0.009 \\
\textit{95\%} &   0.926 $\pm$ 0.005 &   0.896 $\pm$ 0.007 &
&       0.978 $\pm$ 0.005 & 0.973 $\pm$ 0.003 \\
         \bottomrule
    \end{tabular}
    }
    \caption{The amount of data used to create the \textbf{m}achine \textbf{a}nnotator (m.a.) varied between 5\% and 95\% of the training data in each fold. This table compares the downstream F\textsubscript{1} scores of the synthetic and gold-standard NER models. The values are the average F\textsubscript{1} scores and their standard deviation across all five folds.}
    \label{tab:data-4-machine-annotation}
\end{table*}

The experiments in Section \ref{sec:data-4-domain-adaptation} indicated that the synthetic corpora improve when more data are available to domain adapt the model generating the text. However, the values in Table \ref{tab:data-4-domain-adaptation} and the values from the original experiments in Table \ref{tab:data-4-both-steps} differ greatly. The results indicate that using a strong machine annotator -- as in Section \ref{sec:data-4-domain-adaptation} -- explains more of the performance. Another set of experiments was conducted to examine this effect.

In contrast to the previous experiments, these experiments use 95\% of the data for domain adaptation of the generative model that produces the synthetic text. This corpus is still allowed to be four times larger than the original training corpus. The data used to create the machine-annotating NER model is varied between 5\% and 95\%.

Models were trained and evaluated using five-fold cross-validation, and the resulting F\textsubscript{1} scores are listed in Table \ref{tab:data-4-machine-annotation}. The average F\textsubscript{1} scores adhere closely to those of the gold standard model that is trained on real data. This is especially clear when contrasting the scores with what was shown in Table \ref{tab:data-4-domain-adaptation}. This strongly suggests that the performance of the synthetic models is mainly explained by the amount of data available when creating the machine annotators.

\subsection{Using Smaller Generative Models}
\label{sec:model-size}

\begin{table}[ht]
    \centering
    \resizebox{0.45\textwidth}{!}{
    \begin{tabular}{c c c}
    \toprule
    \textbf{Model size} & \textbf{SEPR PHI} & \textbf{MEDDOCAN} \\
    \midrule
         \textit{Small} & 0.883 $\pm$ 0.006 &  0.973 $\pm$ 0.004  \\
         \textit{Larger}  & 0.896 $\pm$ 0.007 & 0.973 $\pm$ 0.003 \\
         \midrule
         \textit{Gold} & \textit{0.926 $\pm$ 0.005} & \textit{0.978 $\pm$ 0.005} \\
    \bottomrule
    \end{tabular}
    }
    \caption{GPT-SW3 and FLOR are available in smaller versions than those used in the other experiments. This table compares the average downstream F\textsubscript{1} scores obtained using the smaller and larger versions for domain adaptation.}
    \label{tab:model-size}
\end{table}

The generative LLMs used for domain adaptation in this study -- GPT-SW3 and FLOR -- are available in different sizes. The previous experiments have used the 6.3 billion and 6.7 billion versions of the models. Although these models are not very large from a research perspective, domain-adapting them still requires expensive hardware. In this experiment, we try synthesizing data using the smaller versions of these LLMs.

Both smaller versions consist of approximately 1.3 billion parameters. Table \ref{tab:model-size} lists the F\textsubscript{1} scores obtained when using 95\% of the data for domain adaptation and for creating the machine annotator. Despite being around five times smaller than their larger counterparts, the smaller models yielded very similar results to their larger counterparts. This suggests that smaller models are a viable alternative, at least for synthesizing data for PII identification. %

\subsection{How Much Synthesis is Enough?}
\label{sec:synthesis-amount}

In all previous experiments, we have exploited the fact that synthetic corpora can be generated indefinitely. This has been represented by letting the corpora be four times larger than the training data. In this experiment, we examine the effect of removing this advantage. In addition to training on the four times larger corpora, we also trained models using corpora of the same size as the training corpora. Finally, we trained models using just 5\% of the synthetic corpora. The data used for domain adaptation and fine-tuning the machine annotator was kept at 95\%.%

\begin{table}[ht]
    \centering
    \resizebox{0.45\textwidth}{!}{
    \begin{tabular}{c c c}
    \toprule
    \textbf{Synthesized} & \multirow{2}{*}{\textbf{SEPR PHI}} & \multirow{2}{*}{\textbf{MEDDOCAN}} \\
    \textbf{amount} & & \\
    \midrule
\textit{5\%} & 0.814 $\pm$ 0.008 & 0.938 $\pm$ 0.006 \\
\textit{100\%} & 0.889 $\pm$ 0.009 & 0.968 $\pm$ 0.005 \\
\textit{400\%} & 0.896 $\pm$ 0.007 & 0.973 $\pm$ 0.003 \\
\midrule
\textit{Gold} & \textit{0.926 $\pm$ 0.005} & \textit{0.978 $\pm$ 0.005} \\
\bottomrule
\end{tabular}
}
    \caption{The synthetic corpora in the other experiments are four times larger than the original gold standards. This table lists the downstream F\textsubscript{1} scores of NER models trained on varying amounts of synthetic data.}%
    \label{tab:synthesis-amount}
\end{table}

Table \ref{tab:synthesis-amount} shows that, for these datasets, generating extra data has a small impact on the results. Generating a synthetic corpus that is the same size as the original corpus yields downstream results that are within one standard deviation of the results from generating a four times larger corpus. This is true both for MEDDOCAN and for SEPR PHI.

\subsection{Diversity of the Generated Data}
\label{sec:diversity-results}

\begin{table*}[ht]
    \centering
        \resizebox{\textwidth}{!}{
    \begin{tabular}{c c c c c c c c}
         \toprule
         \multirow{2}{*}{\textbf{\% for d.a. }} & \multicolumn{3}{c}{\textbf{SEPR PHI}} & & \multicolumn{3}{c}{\textbf{MEDDOCAN}}  \\
         \cmidrule{2-4} \cmidrule{6-8}
         & \textbf{Diversity} & \textbf{Doc. length} & \textbf{Doc. labels} &
         & \textbf{Diversity} & \textbf{Doc. length} & \textbf{Doc. labels} \\
         \midrule
         
\textit{0\%} &  4.28 $\pm$ 0.27 &       53.80 $\pm$ 25.21 &     1.62 $\pm$ 7.25 &
&       2.69 $\pm$ 0.10 &       542.42 $\pm$ 623.43 & 34.46 $\pm$ 111.86 \\
\textit{5\%} &  4.52 $\pm$ 0.30 &       18.28 $\pm$ 16.66 &     0.82 $\pm$ 4.04 &
&       2.58 $\pm$ 0.03 &       644.30 $\pm$ 1729.28 & 57.26 $\pm$ 181.33 \\
\textit{25\%} & 4.27 $\pm$ 0.22 &       16.73 $\pm$ 16.41 &     0.73 $\pm$ 3.95 &
&       2.44 $\pm$ 0.02 &       525.79 $\pm$ 367.61 & 56.34 $\pm$ 27.13 \\
\textit{50\%} & 4.38 $\pm$ 0.15 &       16.41 $\pm$ 16.32 &     0.81 $\pm$ 4.12 &
&       2.37 $\pm$ 0.05 &       519.07 $\pm$ 346.71 & 57.18 $\pm$ 25.18 \\
\textit{95\%} & 4.25 $\pm$ 0.29 &       16.69 $\pm$ 17.57 &     0.85 $\pm$ 4.36 &
&       2.40 $\pm$ 0.03 &       508.63 $\pm$ 347.05 & 57.91 $\pm$ 25.67 \\

         \midrule
         \textit{Gold} & \textit{6.26 $\pm$ 0.03} & \textit{13.12 $\pm$ 18.52} & \textit{0.31 $\pm$ 2.12} &
         & \textit{5.55 $\pm$ 0.03} & \textit{510.28 $\pm$ 427.85} & \textit{48.73 $\pm$ 19.97}              \\
         \bottomrule
    \end{tabular}
    }
    \caption{The average lexical diversity, average length, and average number of annotated labels per document was calculated for the synthetic corpora and for the gold corpora. The synthetic corpora were machine annotated using models trained on 95\% of the data. The \textbf{d}omain \textbf{a}daptation (d.a.) varied between using 95\% of the data, to using none.}
    \label{tab:diversity-results}
\end{table*}

Three different metrics were used to quantify the data themselves. These were lexical diversity, the length of the documents, and the number of entities in the documents. The metrics were calculated both for the generated corpora and for the gold standard corpora. The lexical diversity was estimated by stemming each token and then dividing the number of unique stems by the total number of tokens. Stems were obtained using the Swedish and Spanish Snowball stemmers implemented in NLTK \cite{Bird-Loper-2004}.

Table \ref{tab:diversity-results} lists the three metrics for all of the considered corpora. For the synthetic corpora, the average number of entities was estimated using the strongest machine annotator for each dataset trained using 95\% of the gold corpora. The diversity and average lengths of the synthetic corpora could be calculated before machine annotation.

The lexical diversity of the synthetic data is fairly consistent, regardless of the amount of data used for domain adaptation. It is also consistently lower than in the gold corpora. This is likely due to the temperature being fixed across the configurations. As explained in Section \ref{sec:method-synth}, varying the temperature had a negligible impact on the downstream performance of the synthetic models. However, it is likely that the lexical diversity of the corpora would increase with higher temperatures.

The largest adjustment from adding domain adaptation is that the synthetic corpora become closer to the gold corpora in terms of the number of entities per document and in length. However, the average number of entities per document tends to be noticeably higher in the synthetic corpora than in the gold corpora.

\subsection{Estimating Privacy}
\label{sec:5-grams}
\begin{figure}[t!]
    \centering
    \includegraphics[width=0.7\linewidth]{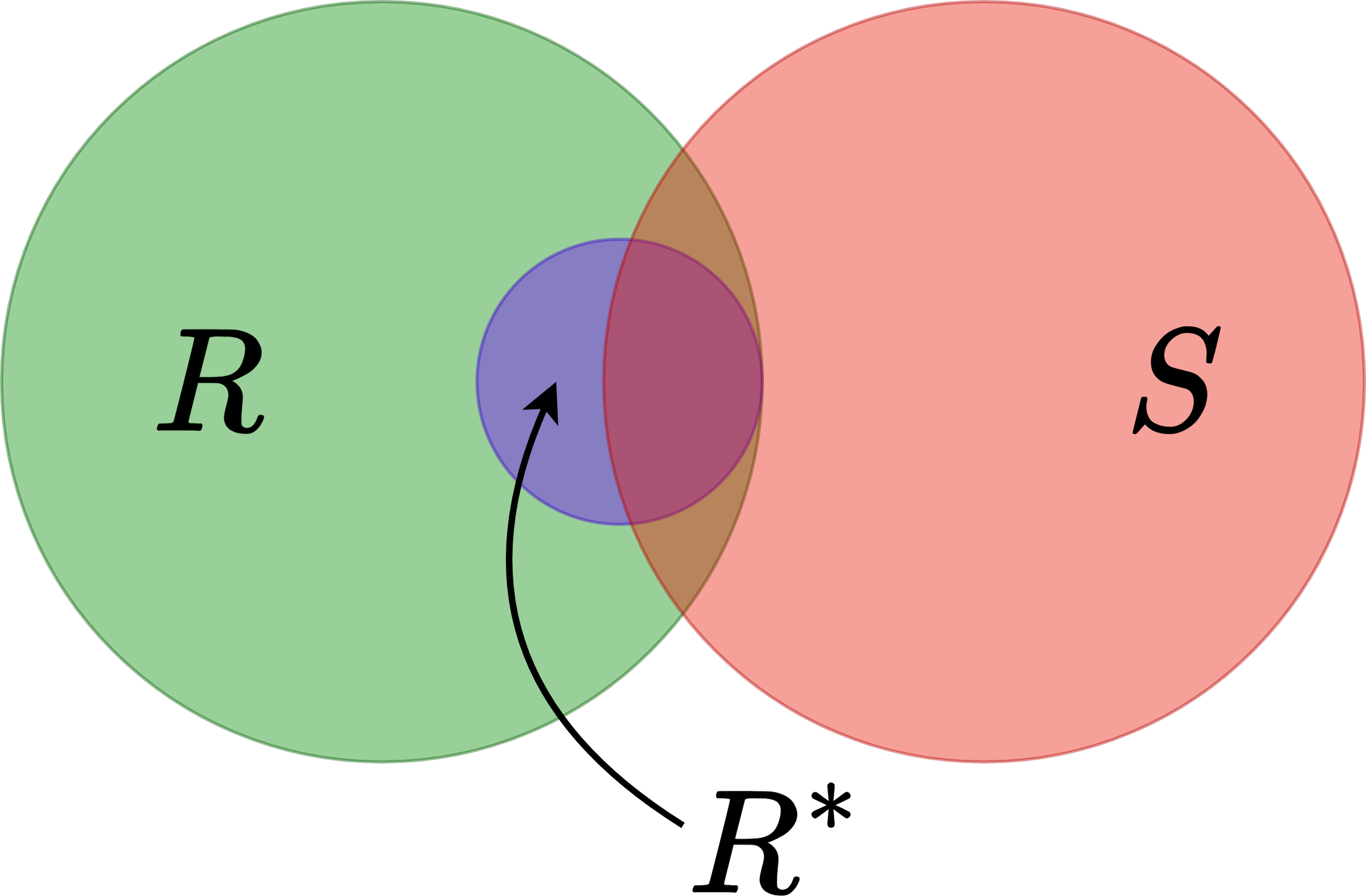}
    \caption{N-gram recall is calculated with all n-grams $R$ in the gold standard corpora, as well as with those in the subset $R^* \subseteq R$ that overlap with sensitive entities. The recall is calculated by counting how many of these are shared with $S$.}
    \label{fig:venn}
\end{figure}

Creating a synthetic variant of a sensitive dataset only protects the original data if the synthetic and sensitive datasets are sufficiently different. A common proxy for measuring these risks is to study the n-grams of the original and synthetic corpora \cite{ive2020generation,hiebel2023can}. %
We calculated the \textit{n-gram recall} of each generated dataset and the training data from which it is derived. This metric measures the proportion of unique n-grams in a reference document that is shared with a candidate document\footnote{It is similar to ROUGE \cite{Lin-2004} but is used on a corpus level rather than for comparing individual documents.}. In this experiment, the reference documents are the real sensitive corpora used for domain adaptation, and the synthetic corpora are the candidates. In other words, given a real corpus with a set of n-grams $R$ and a synthetic corpus with n-grams $S$:
\begin{equation}
    \text{n-gram recall} = \frac{| R \cap S |}{| R |} \; \; .
\label{eq:n-gram-recall}
\end{equation}
Since the data in this study are tagged for PII, a PII-sensitive n-gram recall is also used to estimate the degree of leakage of potentially sensitive information. Instead of considering all n-grams $R$ in the reference document, as in Equation \ref{eq:n-gram-recall}, this metric only considers the n-grams $R^* \subseteq R$ that overlap with sensitive entities in the gold standard corpus. The relation between $R$, $R^*$, and $S$ is illustrated in Figure \ref{fig:venn}.

\begin{table*}[ht]
    \centering

    \resizebox{0.85\textwidth}{!}{
    \begin{tabular}{c c c c c c}
         \toprule
         \multirow{2}{*}{\textbf{\% for d.a. }} & \multicolumn{2}{c}{\textbf{SEPR PHI}} & & \multicolumn{2}{c}{\textbf{MEDDOCAN}}  \\
         \cmidrule{2-3} \cmidrule{5-6}
         & \textbf{All 5-grams} & \textbf{Sensitive 5-grams} &
         & \textbf{All 5-grams} & \textbf{Sensitive 5-grams} \\
         \midrule
\textit{5\%} &    0.328 $\pm$ 0.041 &     0.233 $\pm$ 0.066 &
&       0.005 $\pm$ 0.000 &     0.008 $\pm$ 0.001 \\
\textit{10\%} &   0.216 $\pm$ 0.002 &     0.154 $\pm$ 0.016 &
&       0.003 $\pm$ 0.000 &     0.006 $\pm$ 0.001 \\
\textit{25\%} &   0.183 $\pm$ 0.015 &     0.169 $\pm$ 0.021 &
&       0.003 $\pm$ 0.000 &     0.004 $\pm$ 0.000 \\
\textit{50\%} &   0.134 $\pm$ 0.021 &     0.141 $\pm$ 0.017 &
&       0.002 $\pm$ 0.000 &     0.003 $\pm$ 0.000 \\
\textit{95\%} &   0.122 $\pm$ 0.013 &     0.132 $\pm$ 0.010 &
&       0.002 $\pm$ 0.000 &     0.003 $\pm$ 0.000 \\
\midrule
\textit{0\%} &    0.028 $\pm$ 0.002 &     0.047 $\pm$ 0.002 &
&       0.001 $\pm$ 0.000 &     0.001 $\pm$ 0.000 \\

         \bottomrule
    \end{tabular}
    }

    \caption{5-gram recall values were calculated for each synthetic corpus over five folds. We calculate both the general 5-gram recall and the recall for 5-grams overlapping with PII in the training corpora. The synthetic corpora varied in the amount of data used for \textbf{d}omain \textbf{a}daptation (d.a.) before generation. The bottom row shows the values for the synthetic corpora generated without domain adaptation when compared to the 95\% gold corpora.}%
    \label{tab:5-gram-recall}
\end{table*}

N-gram recall values were calculated for $n = \{3, 5, 10\}$. This was done for each of the five folds in the previous experiments. The n-grams were created by concatenating $n$ tokens from the tokenizers of the domain-adapted generative models. The values for 5-grams are summarized in Table \ref{tab:5-gram-recall} and the other values in Tables \ref{tab:appendix-3-gram} and \ref{tab:appendix-10-gram} in Appendix \ref{appendix:extra-grams}.

All three configurations produced similar patterns. The bottom row of Table \ref{tab:5-gram-recall} shows the average 5-gram recall scores of the synthetic corpora that were generated without any domain adaptation. These values were obtained by comparing these corpora to the 95\% training corpora. %
These values serve as a useful baseline since any shared n-grams in these cases are purely incidental.

More interestingly, the n-gram recall values decrease as more data are used for domain-adapting the generating model. %
Although this trend is not linear -- there is a small increase between 10\% and 25\% for sensitive 5-grams in SEPR PHI -- and the standard deviations are large, there is a clear decrease between using 5\% and using 95\% of the data. The most likely explanation is that the number of unique in-domain n-grams increases as the training data grows, meaning that each individual n-gram is less likely to be memorized. Conversely, using too little data for domain adaptation can cause the model to overfit. %

While both general and sensitive n-gram recall decrease when more data are used, the sensitive n-gram recall is sometimes slightly higher. This could indicate that n-grams overlapping with sensitive entities may be at a higher risk of memorization, although this effect is very small. The n-gram recall values are also significantly higher in SEPR PHI than in MEDDOCAN. There is no clear explanation for this, other than that they differ substantially in their structure. MEDDOCAN contains fewer and longer documents, whereas SEPR PHI contains many but shorter documents. Inspecting the overlapping n-grams also revealed that many of them are related to dates or other vague but oft-occurring categories of PII.

\section{Discussion}

The results in Table \ref{tab:data-4-domain-adaptation} show that domain adaptation -- to a point -- was needed to generate synthetic data of sufficient quality. From a privacy perspective, there is, of course, a risk that domain adaptation causes sensitive data to be memorized and reproduced during data synthesis. The results in Table \ref{tab:5-gram-recall} indicate that a lower proportion of sensitive n-grams were reproduced when more data were used for domain adaptation. On the other hand, using fewer data minimizes the attack surface of the models. If the data requirements are lower, this may make it feasible to, for example, manually de-identify the data, or to audit an automatic de-identification of them.

An example of an application of our results is the cross-institutional validation of NLP systems. In sensitive domains such as the clinical domain, researchers are often barred from sharing their data and trained models due to privacy concerns. A common situation is when research group (A) has created a system that works very well on their in-house data. Due to privacy regulations, another group (B) cannot share their data with (A), and this makes it difficult for (A) and (B) to validate if the system generalizes. An example of an attempt to work under these constraints is described by \citet{Bridal-etal-2022}. They were able to make limited claims of generalization, but were limited due to the restrictions on sharing data. The method for synthesis explored in our paper would allow these research groups to share synthetic versions of their datasets or models, as these are much less sensitive than artifacts based on real data.

Training models using synthetic corpora is safer than using real data, but is no panacea. On the other hand, no currently existing technique for privacy preservation is sufficient when used in isolation. For example, NER-based automatic de-identification covers only a subset of PII \cite{pilan_text_2022}, and differentially private learning for NLP is difficult to implement properly \cite{brown_what_2022,Miranda-etal-2024} and is often inefficient when done so \cite{igamberdiev_dp-nmt_2024}. Synthetic data generation will likely be an important ingredient in many domains to overcome privacy issues.

\section{Conclusions}
Data synthesis is an attractive tool for dealing with data scarcity and privacy risks. However, synthesis itself can be challenging when access to data is constrained.
Through extensive ablation studies -- validated on models and data in two different languages -- our experiments show that not all parts of the synthesis process are equally sensitive to these resource constraints. Domain adapting the models to create high-utility clinical text did not require using all of the data. The experiments show that using between 25\% and 50\% of the data can be enough for domain adaptation, at least for the studied datasets. Furthermore, the experiments also show that using smaller generative LLMs does not necessarily incur a big loss of utility. Instead, our experiments show that the most important factor is the data available to create machine annotations.

These findings are a valuable contribution to the clinical NLP and privacy communities, where sharing real data is often impractical or impossible due to legal constraints. These constraints, while necessary from a privacy perspective, also hinder collaboration. While our studies cannot fully ascertain the privacy of the synthetic corpora, the results indicate that they are less sensitive than the original data. When combined with other safety measures, such as de-identification and secure storage, synthetic data can serve as a basis for collaboration across institutional boundaries.

\section{Limitations}
The results demonstrated the importance of adapting the synthesizing LLM to the clinical domain in order to generate high-utility training data. For synthesizing corpora for PII detection, this was possible to do with very small amounts of data. However, it may be the case that PII detection is a task where the domain-specific details of the data are less important. Indeed, \citet{libbi2021generating} argue that NER data, in general, retain their utility for machine learning even if, qualitatively, their contents lack coherence. In future work, it would be interesting to investigate if the same holds for clinical tasks that are more challenging and domain-specific tasks. %
Nevertheless, PII detection in the clinical domain is an important task in itself and any advances in this area will help to combat the issue of data scarcity.

Our process for synthesis also uses autoregressive language modeling without instruction-tuning. We opted for this design to make the task as simple as possible. As previously mentioned, this may not necessarily be a good design for other types of tasks where the document-level semantics matter more. For example, \citet{kiefer2024instruction} synthesized data for the task of assigning diagnosis codes to discharge summaries by instruction-tuning models to create documents with different characteristics.

Another limitation of the study is the reliance on n-gram-based metrics for estimating privacy risks. This is a common practice \cite{ive2020generation,hiebel2023can} and can detect when data are being reproduced verbatim in the synthetic corpora. On the other hand, n-grams vary greatly in how sensitive they are. We try to address this with our n-gram recall metric that takes PII into account. However, we make limited claims about the privacy of the data and instead focus on their utility.

In Section \ref{sec:model-size}, we find that smaller versions of the generative models could generate data of near-equal utility as their larger counterparts. This was especially clear when generating MEDDOCAN data. An interesting continuation would have been to fine-tune an even smaller LLM using MEDDOCAN data. Unfortunately, the 1.3 billion version of FLOR that we use is the smallest one. GPT-SW3 is available in smaller versions, but proceeding to a monolingual analysis would lower the validity of the results and fell outside the scope of this study. Future work could try similar experiments with languages for which smaller models exist.

\section{Ethical Statement}
This work was conducted under ethical permission no. 2019-05679 granted by the Swedish Ethical Review Authority. MEDDOCAN is a publicly available dataset, where the PII in the documents are unrelated to the original patients. SEPR PHI is available on request and is a manually pseudonymized corpus where all identified PII have been replaced with surrogate values. Because of this, the privacy risks of the experiments in this paper are very small. Regardless, the experiments have been carried out in a computational environment in which only the authors and system administrators have had access to the data. Our experiments are also in accordance with the intended uses of the datasets.

The experiments conducted in this paper required considerable amounts of computational resources. We estimate that creating and evaluating the data and models for our experiments took approximately 130 GPU hours\footnote{The calculations are available in Appendix \ref{appendix:computation}}. Luckily, the experiments were run in Sweden -- where virtually all energy comes from sustainable sources. Nevertheless, the electricity expended when conducting these experiments could have been used for other purposes.

On the other hand, our results indicate that high-utility synthetic corpora can be created using small-scale data and without relying on the very largest LLMs. These results can be particularly helpful for researchers working in resource-constrained environments. This includes not only researchers in, e.g., the clinical domain, but also those working with low-resource languages. These parts of the NLP community are often under-served as increasing focus is placed on terabyte-scale datasets and LLMs with unwieldy amounts of parameters.

Even though synthetic data are safer than sensitive data, there is a risk that other researchers over-interpret our results and use them to justify irresponsible uses of synthetic data. Our focus on constraining the amounts of data used hopefully mediates some of these potential risks. Furthermore, we have thoroughly described the limitations of our results and the scope of our experiments.

\section*{Acknowledgments}
We are grateful for the support for this study from the DataLEASH project and its funder, Digital Futures. The computations were enabled by resources provided by the National Academic Infrastructure for Supercomputing in Sweden (NAISS), partially funded by the Swedish Research Council through grant agreement no. 2022-06725.

\bibliography{custom}

\appendix
\section{Hyperparameters and Software Versions}
\label{appendix:ft-params}
Both GPT-SW3 and FLOR were fine-tuned using the hyperparameters in Table \ref{tab:hyperparams-da}. These were selected based on examples from prior works and through small-scale experiments. SweDeClin-BERT and roberta-base-bne were also trained using the same hyperparameters, listed in Table \ref{tab:hyperparams-ft}. These parameters were also selected based on examples from prior works. If the encoder models got stalled in the local minimum of only predicting the zero class, then training was restarted using the same data and hyperparameters. Due to the many different factors in our synthesis process, a full search of the hyperparameter space was not feasible. The hyperparameters used in the experiments proved sufficiently optimized to obtain high-utility models and clear results.

\begin{table}[ht]
    \centering
    \begin{tabular}{l c}
        \toprule
        \textbf{Parameter} & \textbf{Value} \\
        \midrule
        \textit{r} & 8 \\
        \textit{$\alpha$} & 32 \\
        \textit{Dropout} & 0.05 \\
        \textit{Weight decay} & 0.1 \\
        \textit{Learning rate} & 0.0001 \\
        \textit{Batch size} & 16 \\
        \textit{Epochs} & 6 \\
        \bottomrule
    \end{tabular}
    \caption{The hyperparameters used for domain-adapting the generative LLMs using QLoRA.}
    \label{tab:hyperparams-da}
\end{table}

\begin{table}[ht]
    \centering
    \begin{tabular}{l c}
        \toprule
        \textbf{Parameter} & \textbf{Value} \\
        \midrule
        \textit{Weight decay} & 0.00001 \\
        \textit{Learning rate} & 0.0001 \\
        \textit{Batch size} & 16 \\
        \textit{Epochs} & 6 \\
        \bottomrule
    \end{tabular}
    \caption{The hyperparameters used when fine-tuning the BERT/RoBERTa models for NER.}
    \label{tab:hyperparams-ft}
\end{table}

We also used several Python libraries to implement our experiments. The most important ones are NLTK (version 3.8.1) for stemming when computing diversity, vLLM (version 0.6.1) for synthesis, Axolotl (version 0.6.0) for domain adaptation (version 0.6.0), and Huggingface Transformers (version 4.44.0) for fine-tuning the NER models and for tokenizing the corpora before counting n-grams.

\section{3-Grams and 10-Grams}
\label{appendix:extra-grams}
As mentioned in Section \ref{sec:5-grams}, we calculated the n-gram overlaps for $n = 3, 5, 10$. Results for $n=3$ are listed in Table \ref{tab:appendix-3-gram} and $n=10$ in Table \ref{tab:appendix-10-gram}. The results follow the same overall pattern as in Table \ref{tab:5-gram-recall} but are included for completeness. As expected, 3-grams are much more likely to occur in both corpora and 10-grams a lot \textit{less} likely. Many 3-grams -- due to subword tokenization -- are not full words. In the main part of the paper, we chose to present 5-grams because this struck a balance between what prior studies have used, and the fact that the models we study use sub-word tokenizers.

\begin{table*}[ht!]
    \centering

    \begin{tabular}{c c c c c c}
         \toprule
         \multirow{2}{*}{\textbf{\% for d.a. }} & \multicolumn{2}{c}{\textbf{SEPR PHI}} & & \multicolumn{2}{c}{\textbf{MEDDOCAN}}  \\
         \cmidrule{2-3} \cmidrule{5-6}
         & \textbf{All 3-grams} & \textbf{Sensitive 3-grams} &
         & \textbf{All 3-grams} & \textbf{Sensitive 3-grams} \\
         \midrule
\textit{5\%} &  0.583 $\pm$ 0.040 &     0.540 $\pm$ 0.061 &
&       0.066 $\pm$ 0.003 &     0.072 $\pm$ 0.006 \\
\textit{25\%} & 0.436 $\pm$ 0.019 &     0.426 $\pm$ 0.022 &
&       0.031 $\pm$ 0.000 &     0.039 $\pm$ 0.001 \\
\textit{50\%} & 0.365 $\pm$ 0.028 &     0.358 $\pm$ 0.033 &
&       0.026 $\pm$ 0.000 &     0.037 $\pm$ 0.001 \\
\textit{95\%} & 0.331 $\pm$ 0.018 &     0.321 $\pm$ 0.018 &
&       0.021 $\pm$ 0.000 &     0.034 $\pm$ 0.001 \\
\midrule
\textit{0\%} &  0.180 $\pm$ 0.011 &     0.179 $\pm$ 0.007 &
&       0.019 $\pm$ 0.000 &     0.026 $\pm$ 0.001 \\

         \bottomrule
    \end{tabular}

    \caption{3-gram recall values were calculated for each synthetic corpus. The values are averages and standard deviations over five folds.}
    \label{tab:appendix-3-gram}
    
\end{table*}

\begin{table*}[ht!]
    \centering

    \begin{tabular}{c c c c c c}
         \toprule
         \multirow{2}{*}{\textbf{\% for d.a. }} & \multicolumn{2}{c}{\textbf{SEPR PHI}} & & \multicolumn{2}{c}{\textbf{MEDDOCAN}}  \\
         \cmidrule{2-3} \cmidrule{5-6}
         & \textbf{All 10-grams} & \textbf{Sensitive 10-grams} &
         & \textbf{All 10-grams} & \textbf{Sensitive 10-grams} \\
         \midrule
\textit{5\%} &   0.294 $\pm$ 0.019 &     0.040 $\pm$ 0.016 &
&       0.000 $\pm$ 0.000 &     0.000 $\pm$ 0.000 \\
\textit{25\%} &  0.137 $\pm$ 0.005 &     0.033 $\pm$ 0.008 &
&       0.000 $\pm$ 0.000 &     0.000 $\pm$ 0.000 \\
\textit{50\%} &  0.082 $\pm$ 0.008 &     0.022 $\pm$ 0.006 &
&       0.000 $\pm$ 0.000 &     0.000 $\pm$ 0.000 \\
\textit{95\%} &  0.052 $\pm$ 0.007 &     0.022 $\pm$ 0.004 &
&       0.000 $\pm$ 0.000 &     0.000 $\pm$ 0.000 \\
\midrule
\textit{0\%} &   0.001 $\pm$ 0.000 &     0.002 $\pm$ 0.000 &
&       0.000 $\pm$ 0.000 &     0.000 $\pm$ 0.000 \\

         \bottomrule
    \end{tabular}

    \caption{10-gram recall values were calculated for each synthetic corpus. The values are averages and standard deviations over five folds.}
    \label{tab:appendix-10-gram}
\end{table*}

\section{Computational Requirements}
\label{appendix:computation}
The experiments in this paper consumed many GPU hours due to the large number of configurations required for the ablation study. The GPUs were provided by the National Academic Infrastructure for Supercomputing in Sweden. Unfortunately, the environment offers no straightforward way of computing the GPU hours. In this appendix, we estimate the GPU hours required to run the experiments based on data from the logs.

Domain-adapting the FLOR 6.3B model and synthesizing the corpora, for all amounts of domain-adaptation considered in this paper, took 7.5 hours. Domain-adapting GPT-SW3 took 10 hours. Both these processes used four \textit{Nvidia A100} GPUs and a total of 70 GPU hours.

The encoder NER models were trained using single \textit{Nvidia V100} GPUs. Training the SweDeClin-BERT models on one 95\% portion of the data took approximately 6 minutes. Training roberta-base-bne on 95\% of a MEDDOCAN fold took approximately 5 minutes. Based on other logs for other configurations, the time scales linearly with the amount of data. Based on this assumption, the models presented in this paper took an additional 60 GPU hours to train.

\end{document}